\documentclass[letterpaper, 10 pt, conference]{ieeeconf}  

\IEEEoverridecommandlockouts                              

\usepackage{ifpdf}
\usepackage{graphicx}
\ifpdf
\usepackage{amsmath}
\usepackage{autobreak}
\usepackage{hhline}
\usepackage{multirow}
\usepackage{hyperref}
\usepackage{graphicx}

\usepackage{algorithmic}
\usepackage{algorithm}  
\usepackage{dblfloatfix}
\usepackage{blindtext}

\usepackage{amsmath}

\graphicspath{{Figs/}{Figs/PDF/}}
\else
\graphicspath{{Figs/}}
\fi
\usepackage{hhline}
\usepackage{color}
\usepackage{calc}
\let\labelindent\relax
\usepackage{enumitem}
\usepackage{bm,amsmath}

\usepackage{nomencl}
\usepackage{multirow}

\usepackage{url}
\usepackage{nomencl}
\usepackage{multirow}
\usepackage{multicol}
\usepackage{multirow}

\usepackage{gensymb}
\usepackage{bbding}
\usepackage{booktabs}
\usepackage{makecell}
\hypersetup{
	colorlinks = true,
	linkcolor = blue,
	citecolor = blue
}
\usepackage{subcaption}

\usepackage{xcolor}

\usepackage{caption}
\usepackage[flushleft]{threeparttable}
\usepackage{cite}
\usepackage{times}
\usepackage{epsfig}
\usepackage{graphicx}
\usepackage{amsmath}
\usepackage{amssymb}

\usepackage[maxfloats=256]{morefloats}
\title{\LARGE \bf
Vis2Hap: Vision-based Haptic Rendering by Cross-modal Generation
}

\author{Guanqun Cao$^{1}$, Jiaqi Jiang$^{ 2}$, Ningtao Mao$^{3}$, Danushka Bollegala$^{1}$, Min Li$^{4}$, and Shan Luo$^{2}$
\thanks{This work was funded in part by the EPSRC project ``ViTac: Visual-Tactile Synergy for Handling Flexible Materials'' (EP/T033517/2).}
\thanks{$^{1}$G. Cao and D. Bollegala are with the Department of Computer Science, University of Liverpool, Liverpool L69 3BX, United Kingdom. Emails:~\{\tt\small g.cao, danushka\}@liverpool.ac.uk.}
\thanks{$^{2}$J. Jiang and S. Luo are with the Department of Engineering, King's College London, London WC2R 2LS, United Kingdom. E-mails: {\tt\small \{jiaqi.1.jiang, shan.luo\}@kcl.ac.uk}.}
\thanks{$^{3}$N. Mao is with School of Design, University of Leeds, LS2 9JT, United Kingdom. E-mail: {\tt\small n.mao@leeds.ac.uk}.}
\thanks{$^{4}$M. Li is with School of Mechanical Engineering, Xi'an Jiaotong University, Xi'an 710049, China. E-mail: {\tt\small min.li@mail.xjtu.edu.cn}.}
}
\begin{document}
\maketitle

\begin{abstract}
To assist robots in teleoperation tasks, haptic rendering which allows human operators access a virtual touch feeling has been developed in recent years. 
Most previous haptic rendering methods strongly rely on data collected by tactile sensors. 
However, tactile data is not widely available for robots due to their limited reachable space and the restrictions of tactile sensors.
To eliminate the need for tactile data, in this paper we propose a novel method named as \textit{Vis2Hap} to generate haptic rendering from visual inputs that can be obtained from a distance without physical interaction. We take the surface texture of objects as key cues to be conveyed to the human operator. To this end, a generative model is designed to simulate the roughness and slipperiness of the object's surface. To embed haptic cues in Vis2Hap, we use height maps from tactile sensors and spectrograms from friction coefficients as the intermediate outputs of the generative model. Once Vis2Hap is trained, it can be used to generate height maps and spectrograms of new surface textures, from which a friction image can be obtained and displayed on a haptic display. The user study demonstrates that our proposed Vis2Hap method enables users to access a realistic haptic feeling similar to that of physical objects. The proposed vision-based haptic rendering has the potential to enhance human operators' perception of the remote environment and facilitate robotic manipulation.


\end{abstract}


\section{INTRODUCTION}


Haptic rendering has been developed to assist robots in teleoperation tasks recently, which allows human operators access to a virtual touch feeling.
Various kinds of hardware devices have been developed to provide humans haptic feedback based on different working principles, such as vibrotactile feedback~\cite{liu2020toward}, electrovibration feedback~\cite{bau2010teslatouch} and thermal feedback~\cite{benali2003thermal}.
As one kind of haptic-rendering devices, electrovibration-based haptic displays, which allow users to feel frictional force changes as they move their fingers over the display, have the potential to simulate surface characteristics of objects, e.g., frictional information, roughness, and textures~\cite{klatzky2019detection,icsleyen2019tactile,jiao2018data}.

\begin{figure}[tbp]
	\centering
	\includegraphics[width=0.8\linewidth]{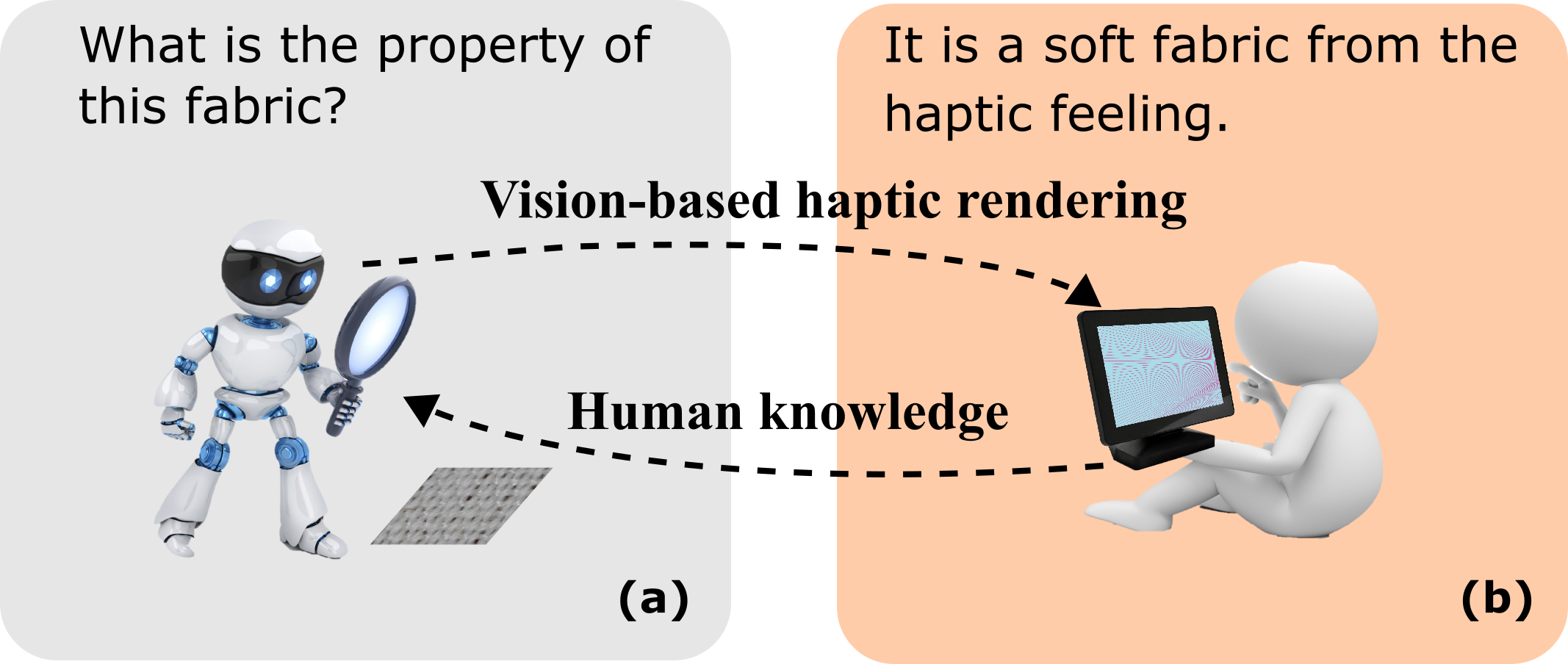}
    \caption{Haptic rendering from vision enables human operators to assist robots to understand the properties of objects and further facilitate robotic manipulation. \textbf{(a)}: A robot has a glance at the fabric but still is unsure about its properties. \textbf{(b)}: A human operator can feel the fabric's textures via sliding her/his finger on the haptic rendering screen, where a virtual fabric is displayed.}
	\label{fig:bigpicture}
	\vspace{-2.0em}
\end{figure}

Most previous haptic rendering methods are limited to reproducing the properties of an object's surface by using tactile signals.
To collect tactile signals, the teleoperation robot is required to contact the target object physically.
However, a robot's attempts to reach every object are inefficient and time-consuming.
Additionally, the target object could also be inaccessible if an obstacle blocks the moving path of the robot. 
Therefore, it is desirable to provide human operators with haptic rendering without requiring tactile signals.

To address the above problem, we propose a haptic rendering framework \textit{Vis2Hap} based on a cross-modal generation model that uses visual images, which can be obtained from a distance, to generate the signals for haptic rendering.
As shown in Fig.~\ref{fig:bigpicture}, vision-based haptic rendering methods allow human operators to assist robots to understand the physical properties of objects without physical contact and tactile data, and help to facilitate robotic manipulation based on human knowledge.

Specifically, the surface textures of objects are taken as key cues to be conveyed in our haptic rendering.
The touch feeling of the surface texture largely depends on two aspects, i.e., \textbf{\textit{roughness}} and \textbf{\textit{slipperiness}}~\cite{mao2016smoothness}.
In our Vis2Hap, the height maps of object's surface, which demonstrate the height changes at high frequencies of the object's surfaces, and spectrograms of dynamic friction coefficients, which can assess the slipperiness, are generated from vision as the intermediate outputs. 
Then, we combine the generated signals to obtain friction images as the input to the haptic display for haptic rendering.
The evaluation results from users demonstrate that our proposed haptic rendering from vision  has a high similarity with the touch feeling of the physical object's surface.

In summary, the contributions of this paper are three-fold:

\begin{enumerate}

    \item We develop a haptic rendering framework named \textit{Vis2Hap} that generates height and frictional information from visual inputs for haptic rendering, for the first time;
    
     \item The generated height maps and friction coefficients, demonstrating the roughness and slipperiness of the object's surface respectively, are combined together to improve the realism of haptic rendering;
    
    \item{A set of experiments demonstrate our proposed Vis2Hap method is capable of providing a realistic haptic feeling similar to that of physical objects without using tactile signals, which enables human operators to assist robots in understanding objects in the remote environment. }

\end{enumerate}


\section{Related Works}

\subsection{Cross-modal Visual-tactile Generation}

Recently, cross-modal visual-tactile generation has made progressive research and attracted a lot of researchers. Lee \textit{et al.}~\cite{lee2019touching} proposed a cross-modal data generation framework based on conditional Generative Adversarial Network (cGAN) to generate pseudo tactile textures from visual images, using the data collected from fabrics.
Cai \textit{et al.}~\cite{cai2021visual} proposed a residue-fusion module based on the generative model for cross-modal generation between visual images and accelerometer signals.
Li \textit{et al.}~\cite{li2019connecting} adopted the generative model to perform two prediction tasks: generating tactile signals from visual videos; reconstructing a visual scene that indicates which object is touched from a tactile input.  
Zhang \textit{et al.}~\cite{zhang2020generative} proposed a generative partial visual-tactile fused framework for clustering where the generated data is used to mitigate the missing data.
However, in these works tactile data was generated from visual data, and no prior works attempted to generate haptic rendering to provide haptic feedback to a human operator as in our work.

\subsection{Haptic rendering based on visual input with an electrovibration haptic display}


Several studies have been conducted on providing haptic rendering from visual information, e.g., using shadings, shapes, and gradients of visual textures.
{\.I}{\c{s}}leyen \textit{et al.}~\cite{icsleyen2019tactile} investigated how the roughness experience changes against different spatial periods and normal force according to the shape of virtual gatings on an electrovibration haptic display.
Wang \textit{et al.}~\cite{wang2014electrostatic} developed a tactile-rendering method to obtain the height information by implementing shape from shading with Gaussian bump.  
Wu \textit{et al.}~\cite{wu2017tactile} proposed a mapping model to get frequency and amplitude based on the gradients in visual textures, which is able to demonstrate the hardness and height on the electrovibration-based haptic display.
However, these methods only provided a limited haptic feeling, e.g., a single value of roughness, frequency, amplitude or height. In contrast, surface textures are rendered on a haptic display in our work.

\subsection{Haptic rendering based on tactile signals with an electrovibration haptic display}
Another popular method is to employ the tactile sensor to record the tactile data of the contacting surface and reproduce the haptic feeling using the collected tactile data.  
Jiao \textit{et al.}~\cite{jiao2018data} measured the friction coefficients from the collected frictional and normal forces and replay them on the haptic display by controlling the voltage to the display.
Ilkhani \textit{et al.}~\cite{ilkhani2017data} proposed a texture rendering algorithm to reproduce the acceleration signal on the haptic display, and a comparison is conducted between simulated feeling and real objects.
Zhao \textit{et al.}~\cite{zhao2019design} combined the acceleration signals and friction properties to improve the haptic rendering.

To eliminate the complexity in tactile data collection, Cai \textit{et al.}~\cite{cai2022gan} proposed a generative model to synthesise the frictional signals from visual images, which are then rendered on a haptic display.
However, friction coefficients were only considered in a straight line and the height disparities over the object's surface were ignored.
In this work, we use visual information and generative models to generate signals that reflect both roughness and slipperiness, and combine them together for haptic rendering, for the first time.

\section{Methodologies}
\begin{figure*}[t]
	\centering
	\includegraphics[width=0.85\linewidth]{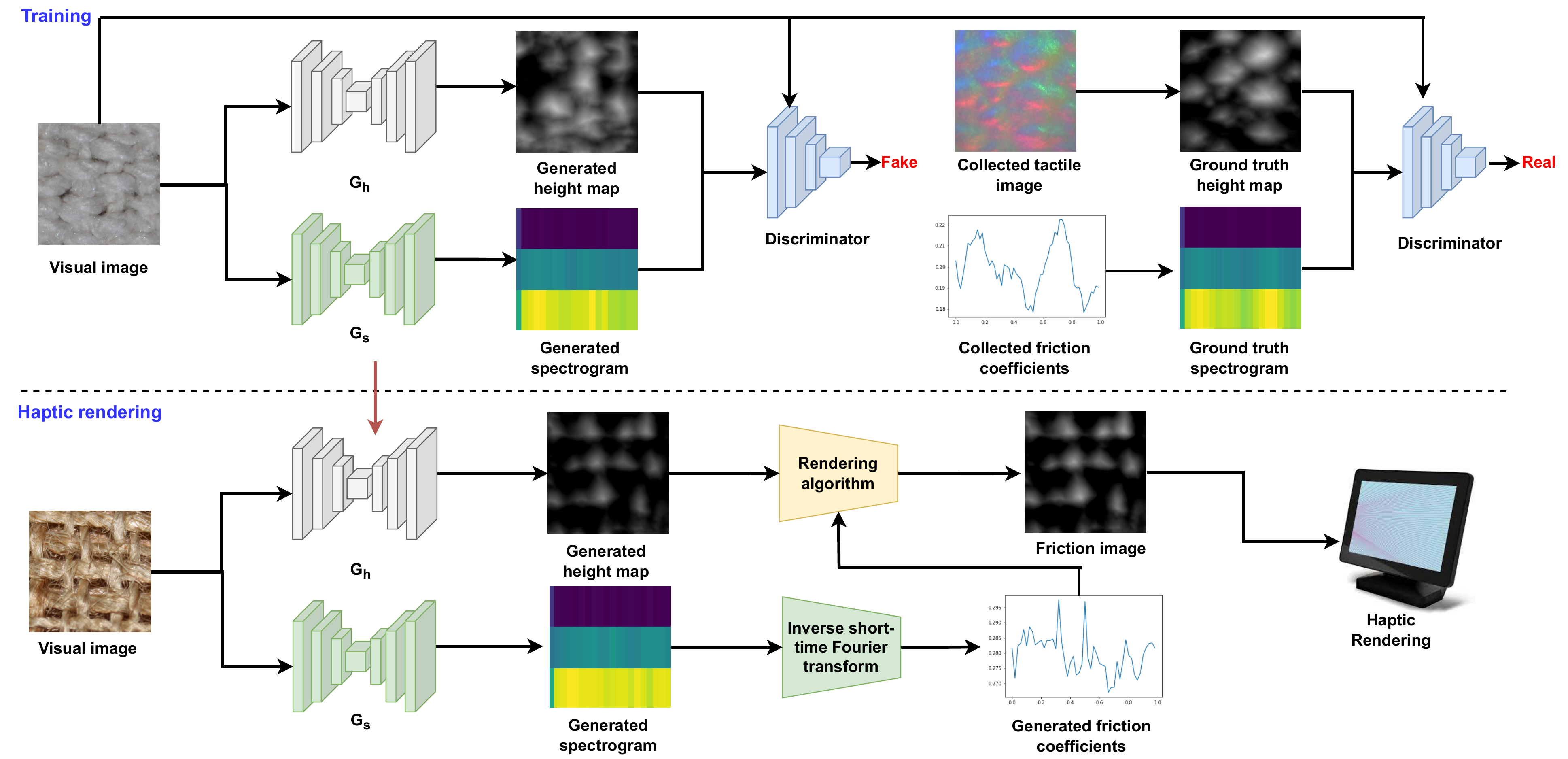}
	\caption{\textbf{The proposed \textit{Vis2Hap} haptic rendering framework.} Two generators $G_h$ and $G_s$ are implemented to generate the height maps of the object's surface and spectrogram of friction coefficients, respectively. After training, the generators are capable of generating realistic signals from corresponding visual images for haptic rendering. In the haptic rendering, the spectrogram is transformed to the wave-format friction coefficients using the inverse short-time Fourier transform algorithm, which are then used to scale the generated height map to produce a friction image for haptic rendering. }
	\label{fig:framework}
	\vspace{-2.0em}
\end{figure*}

As shown in Fig.~\ref{fig:framework}, we develop our Vis2Hap based on a conditional Generative Adversarial Network (cGAN)~\cite{mirza2014conditional} that generates tactile signals of friction coefficients and height maps of the object's surface from visual images as intermediate outputs.  The obtained signals are then utilised to create friction images for haptic rendering. 

\subsection{Height maps and friction coefficients generation}
\begin{figure}[t]
	\centering
    \includegraphics[width=0.85\linewidth]{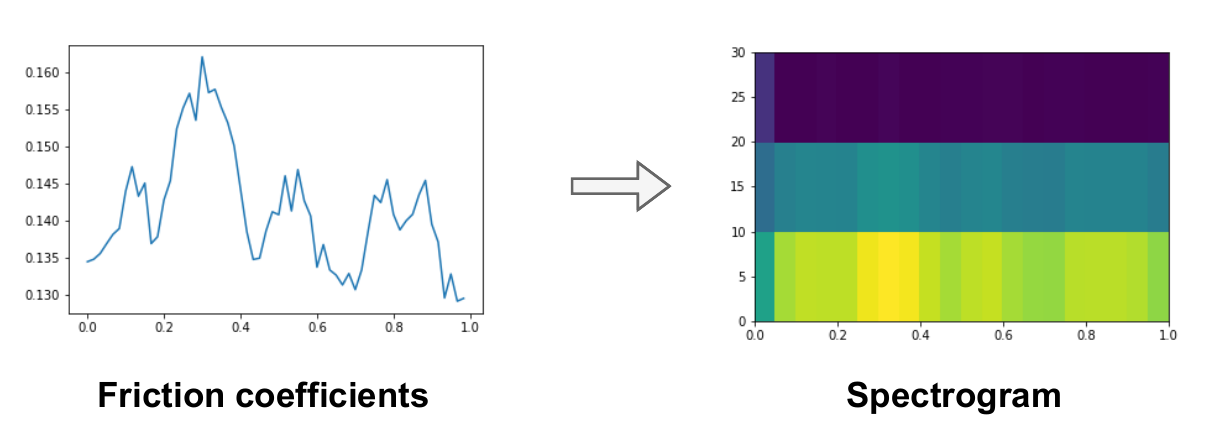}
	\caption{\textbf{Spectrogram.} 
	The collected friction coefficients are transformed into spectrograms by using STFT. In the friction coefficients, the x-axis and y-axis represent time and the value of coefficients, respectively. In the spectrogram, the x-axis and y-axis represent time and frequency, respectively, with the colour representing the amplitude (a brighter colour denotes a higher amplitude).  }
	\label{fig:transform2}
	\vspace{-0.5em}
\end{figure}

\begin{figure}[t]
	\centering
    \includegraphics[width=0.85\linewidth]{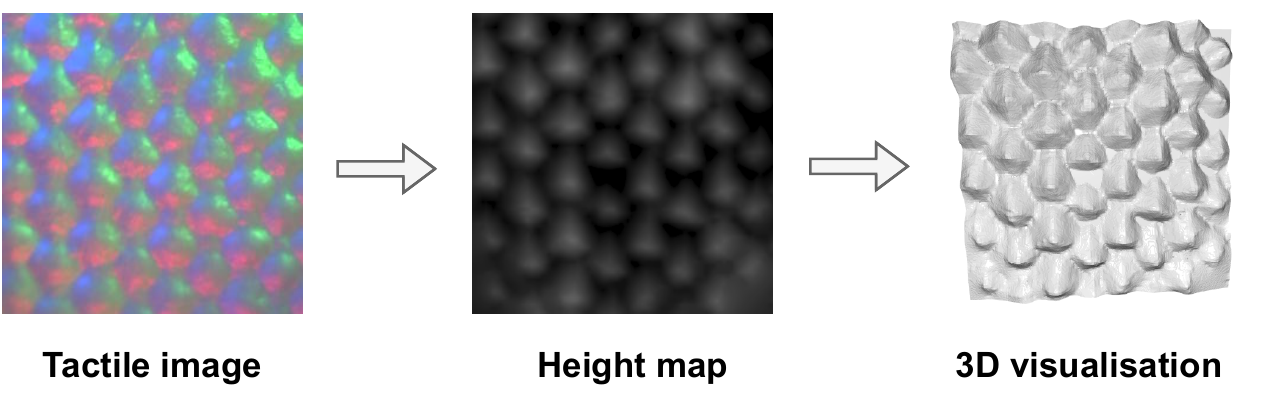}
	\caption{\textbf{Height map.} Height information is reconstructed from tactile images by using a photometric stereo algorithm. The brighter location indicates a higher height.}
	\label{fig:transform1}
	\vspace{-2.0em}
\end{figure}

Our generative model consists of two generators $G_s$ and $G_h$ as well as a discriminator $D$, as shown in the training phase of Fig.~\ref{fig:framework}.
The visual images $\{x_i\}_{i=1}^N$, tactile images $\{t_i\}_{i=1}^N$ and friction coefficients data $\{f_i\}_{i=1}^N$ are used to train the generative model.
Specifically, friction coefficients data over different locations, considered as time series, is converted into 2D spectrograms $\{f_i\}_{i=1}^N \rightarrow \{s_i\}_{i=1}^N$, which can illustrate the pattern of coefficients change in time-frequency domain effectively, by using Short-Time Fourier Transform (STFT)~\cite{sejdic2009time} (as shown in Fig.~\ref{fig:transform2}).
The height maps of the object's surface are reconstructed from tactile images $\{t_i\}_{i=1}^N \rightarrow \{h_i\}_{i=1}^N$ by using a photometric stereo algorithm~\cite{yuan2017gelsight}, as shown in Fig.~\ref{fig:transform1}, to provide the ground truth for training the generator.

During the training process, the generators $G_s$ and $G_h$ take visual images $x$ to generate corresponding spectrograms of friction coefficients and height maps of the object's surfaces, respectively.
The discriminator $D$ uses the input visual image $x$ as auxiliary information, along with the generated results as well as the height maps and spectrograms from the real distribution, to train the model to identify whether the input to $D$ is from a real distribution or a generated distribution.

For the training of the generative model, we optimise the generators and discriminators iteratively.
Concretely, the discriminator $D$ is trained by minimising:
\begin{equation}
\begin{aligned}
\mathcal{L}_{D} (D)=& -\mathbb{E}_{x, s, h}[\log D(x, s, h)]\\
& -\mathbb{E}_{x}[\log (1-D(x, G_s(x), G_h(x))],
\end{aligned}
\end{equation}
At the same time, the generators are trained to generate synthetic height maps and spectrograms to fool the discriminator by minimising:
\begin{equation}
\begin{aligned}
\mathcal{L}_{G}(G_s, G_h)= -\mathbb{E}_{x}[\log (D(x, G_s(x), G_h(x))].
\end{aligned}
\end{equation}
Through the competition, the generators are capable to generate realistic spectrograms and height maps for haptic rendering.
Moreover, we minimise the L1 distance between the generated data and real data for less blurring~\cite{isola2017image}:
\begin{equation}
\begin{aligned}
\mathcal{L}_{L 1}(G_s,G_h)&= \mathbb{E}_{x, s}\left[\|s-G_s(x)\|_{1}\right]\\
&+\mathbb{E}_{x, h}\left[\|h-G_h(x)\|_{1}\right].
\end{aligned}
\end{equation}

\subsection{Haptic rendering algorithm}
A haptic rendering algorithm is designed for the electrovibration haptic display that relies on bitmapped stimulus images (friction images) as input, where the location on the screen with a higher pixel value input produces higher friction and the place with lower pixel value input produces less friction. 

In the physical world, when we use a finger to slide over the object's surface, the surface of the finger is inserted into the textures of the object due to the pressure. As a result, the locations with higher heights prevent the finger from moving and the locations with lower heights  provide less friction~\cite{lafaye2006ploughing}.
However, haptic feelings of the object's surfaces can vary greatly according to their different properties in slipperiness, even with the same texture.
To this end, we can use the friction coefficients to scale the value of height map accordingly as friction images for haptic rendering.

Concretely, as shown in the test phase of Fig.~\ref{fig:framework}, the trained generators $G_h$ and $G_s$ are used to generate the height maps $h^{\prime}= G_{h}(x^{\prime})$ and spectrograms $s^{\prime}= G_{s}(x^{\prime})$ of test objects, respectively, where $x^{\prime}$ is the visual images of test objects.
Then, the spectrograms are converted to the wave-format friction coefficients $f^{\prime}= \mathrm{istft}(s^{\prime})$ by using the inverse short-time Fourier transform algorithm~\cite{yang2008study}.
Consequently, the scaled height maps can be denoted as:

\begin{equation}
\begin{aligned}
m^{\prime}= f_{avg}^{\prime} * h^{\prime},
\end{aligned}
\end{equation}
where $f_{avg}^{\prime}$ denotes the average value of friction coefficients over different locations.
Finally, we map the scaled height maps to friction images for haptic rendering according to the range of the input values of haptic display: 
\begin{equation}
\begin{aligned}
m_{{i,j}}^{n}&=(\max(p)-\min(p))*\frac{{m_{i,j}^{\prime}}^{n}-\min (m^{\prime})}{\max (m^{\prime})-\min (m^{\prime})}\\
&+\min(p),
\end{aligned}
\end{equation}
where $n$ represents the index of the test objects, and $i, j$ denotes the location of pixels. $p$ denotes the  range of input pixel's value of haptic display, which is from 0-255.

\section{Data Collection}

\begin{figure}[t]
	\centering
	\includegraphics[ width=0.85\linewidth]{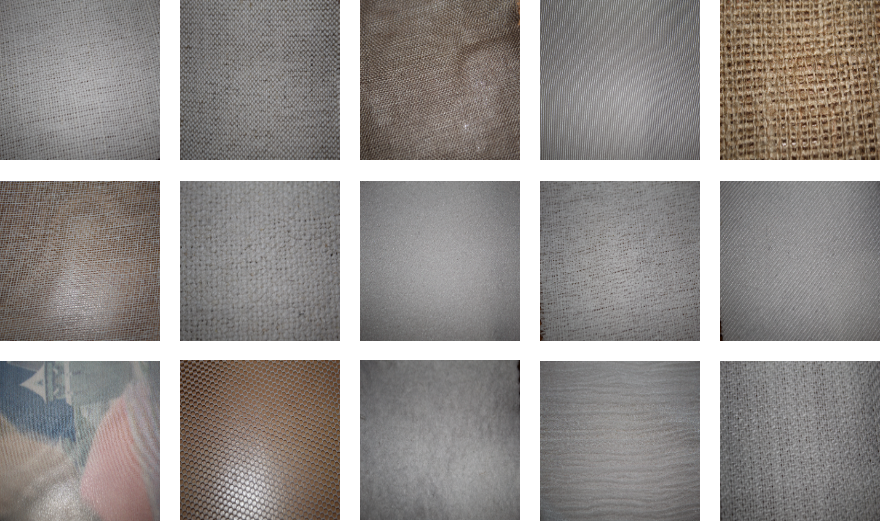}
	\caption{\textbf{Physical fabrics}. There are 15 kinds of fabrics used in our experiments, which are made of different materials and manufactured using different weaving or knitting methods. }
	\label{fig:fabrics}
	\vspace{-0.5em}
\end{figure}

As shown in Fig.~\ref{fig:fabrics}, a total of 15 kinds of fabrics are selected in our experiments, which are made of different materials and manufactured using different weaving or knitting techniques, e.g., tarlatan cotton, loomstate, and zeddana silk.
Although most of the selected fabrics are of the same colour (white colour), their various surface textures give each fabric a unique appearance. 
Furthermore, their surface height distribution and slipperiness characteristics are different from each other, resulting in a variety of haptic feelings.

A weakly paired dataset is collected by sampling from them, including visual images, height maps, and spectrograms of friction coefficients.

\begin{figure}[t]
	\centering
    \includegraphics[width=0.85\linewidth]{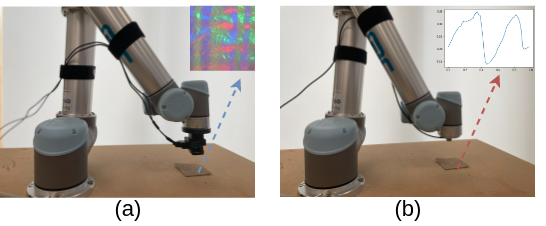}
	\caption{\textbf{Data collection.} (a) a GelSight sensor is used to collect tactile images from fabrics.
	(b) a force/torque sensor Nano17 is used to collect the friction coefficients data from fabrics. }
	\label{fig:datacollection}
	\vspace{-2.0em}
\end{figure}

\noindent \textbf{Visual images.} The visual images of fabrics are collected with a digital camera Canon 2000D. 
Fabrics are placed on a flat plane with the image plane approximately parallel to them.
For each piece of fabric, 5 colour images are taken under different in-plane rotations.
Moreover, data augmentation is performed such as random rotation, flip and Gaussian noise.
As a result, there are 3,375 colour images of fabrics in total in our data set.


\noindent \textbf{Height maps from tactile images.} As shown in Fig.~\ref{fig:datacollection} (a), a camara-based tactile sensor GelSight sensor~\cite{yuan2017gelsight} is mounted on the UR5 robot arm to press against the flat fabrics by a constant force (around 20N) to collect tactile images by sampling from different locations.
After collecting tactile images, following~\cite{yuan2017gelsight}, we use the photometric stereo algorithm to reconstruct the height map that demonstrates the vertical displacement on the sensor.
Consequently, there are 3,375 height maps of the surface textures in the dataset.

\noindent \textbf{Spectrograms of friction coefficients.} 
Apart from the height maps, we collect friction coefficients on a straight line of fabrics to measure the slipperiness of each fabric.
Specifically, the UR5 robot arm is equipped with a force/torque sensor Nano17, with a sampling rate of around 60Hz, to move over the fabrics (as shown in Fig.~\ref{fig:datacollection} (b)).
The sensor is controlled to slide along the fabric for 4 cm at a steady speed of 5 mm/s after being pressed against fabrics with a force of around 15 N.
By using the collected friction and the normal pressure force, we can calculate the coefficients of friction over a straight line for each fabric.
Then, we apply the short-time Fourier transform (STFT)~\cite{sejdic2009time} to convert the friction coefficients into spectrograms.
Finally, we have 3,375 spectrograms after subsampling on friction coefficients. 

We randomly split the whole dataset with a ratio of 8:1:1 for training, validation and testing, respectively.
\section{Experimental Setup}

\subsection{Haptic display }
A TanvasTouch Desktop Development Kit\footnote{https://tanvas.co/products/tanvastouch-dev-kit} is used for haptic rendering.
The Tanvas haptic display, based on electrovibration mechanism, is able to provide software-defined haptics through the SDK.
The Tanvas haptic display has a 10.1-inch screen with a resolution of 1280 $\times$ 800 pixels.
The haptics are mapped 1:1 to the input friction image.
The value of pixels of the friction image ranges from 0 to 255, where 0 represents the friction that naturally exists on the surface of the haptic display, and 255 represents the highest amount of friction that the device is capable of producing. 
The device will output the required interaction as soon as the finger is over a location where a friction image has been added. 

\subsection{Baselines of haptic rendering of virtual textures}
To evaluate the effectiveness of our proposed method, a number of baseline methods that employ different input signals are used for comparison. 
The different inputs are listed below:
\begin{enumerate}
    \item Grey-scale visual images (denoted as $v_{grey}$) that are obtained by the weighted mean of RGB channels of colour visual images;
    \item Shape from shading (denoted as $v_{shape}$) using visual images~\cite{wang2014electrostatic};
    \item Generated friction coefficients (denoted as $f_{g}$)~\cite{cai2022gan};
    \item Generated height maps (denoted as $h_{g}$); 
    \item Generated height maps $h_{g}$ \& generated friction coefficients $f_{g}$ (input of proposed method);
    \item Grey-scale tactile images (denoted as $t_{grey}$);
    \item Ground truth friction coefficients (denoted as $f$)~\cite{jiao2018data};
    \item Ground truth height maps (denoted as $h$);
    \item Ground truth height maps $h$ \& ground truth friction coefficients $f$.
\end{enumerate}


\subsection{Experimental setup for user study}

\begin{figure}
	\centering
	\includegraphics[trim=0 60 0 12, clip,width=0.78\columnwidth]{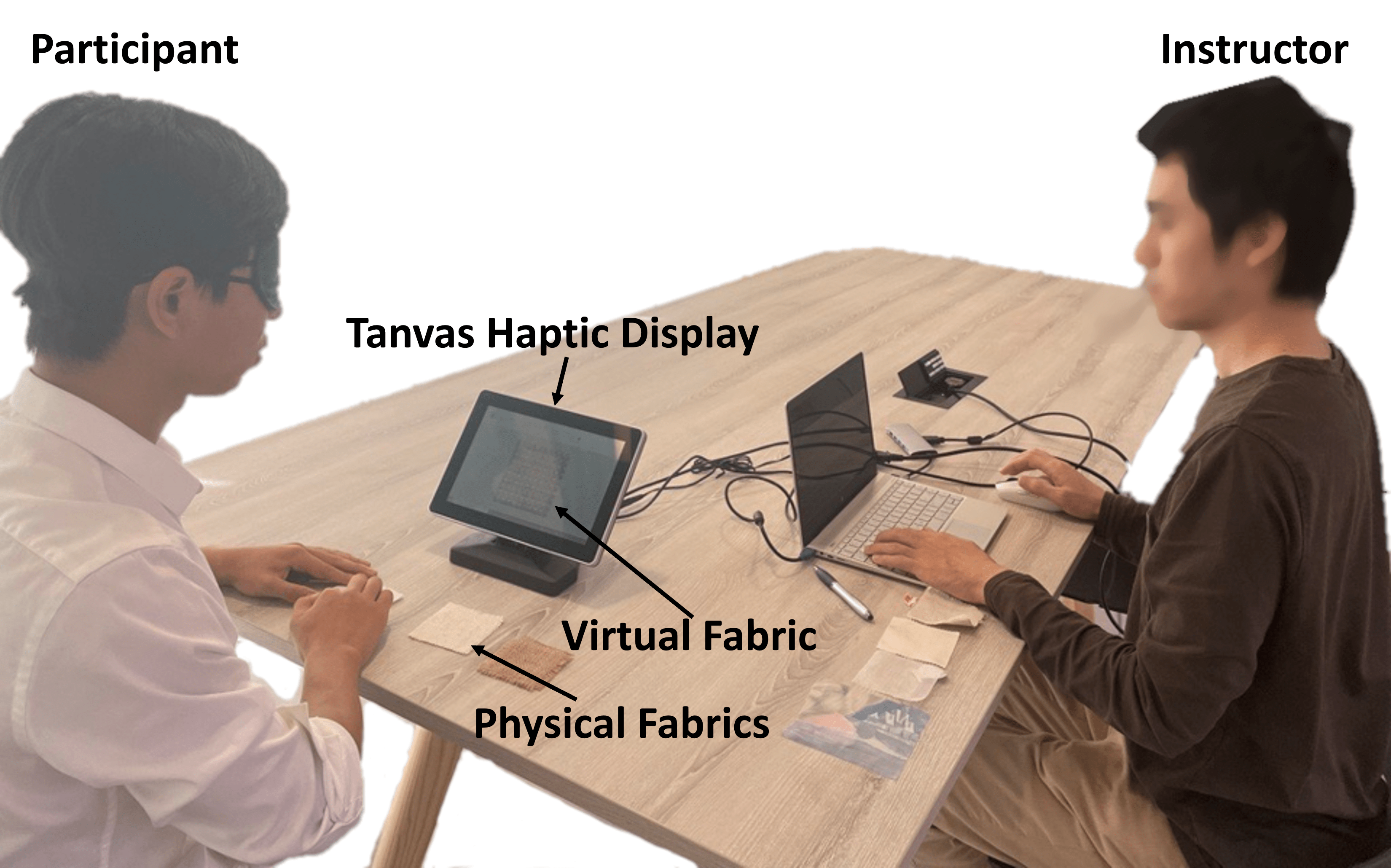}
	\caption{\textbf{Experimental setup.} The participant is blinded with an eye mask and let to touch the physical fabrics on the table and the virtual fabrics on the haptic display. The instructor will change the physical and virtual fabrics and record the reaction from the participant. }
	\label{fig:setup}
	\vspace{-1.5em}
\end{figure}

In our experiment, we investigate if the haptic rendering based on our proposed methods can have a high similarity of haptic feedback to human subjects with the  physical fabrics. 
We recruit 10 volunteers (8 males and 2 females) and the age of participants ranges from 24 to 31.
None of them has experience with haptic displays.
To reduce the time consumption of the testing, 7 pieces of fabrics are selected randomly in our user study.
As illustrated in Fig~\ref{fig:setup}, to study the haptic rendering only in this work without the effect of visual cues (we would like to study the fusion of visual cues and haptic rendering in a future work), the participants will be blinded to touch the physical fabrics and the haptic display, and then be asked to respond to a series of designed questions as described in Table~\ref{question1}. 
Before the experiments, the haptic rendering of random fabrics will be given on the haptic display and let participants have a mock-up test and be familiar with the device.

For Q1, the participants will be given one haptic rendering on the haptic display and three physical fabrics, and the physical fabric corresponding to haptic rendering is among these three physical fabrics and the other two are randomly selected.  
The participants will be asked to match the haptic rendering with the most similar physical fabrics through haptic feeling.
After testing Q1 for all testing fabrics, questions Q2-Q4 will be asked.
Each physical fabric and its corresponding haptic rendering will be shown to participants one by one.
Specifically, a Haptic Analog Scale (HAS) rating based on a Visual Analog Scale (VAS)~\cite{lee1991validity} rating is proposed to measure the degree of similarity between physical fabrics and haptic rendering:
in the experiment, participants are required to memorise a nine-sectioned line segment before the test and grade similarity on this analog scale during the testing;
A rating of 0 indicates that the haptic rendering and the haptic feeling of physical fabric are unrelated, and a rating of 10 indicates that the haptic rendering is very similar to the properties of physical fabrics. 
To help participants understand the scale, examples are provided to them as a reference: the haptics of a piece of sandpaper and a piece of silk are unrelated, which receives a score of 0; the haptics from two same fabrics are totally the same, which receives a score of 10. 

\begin{table}
	\centering
		\caption{The participants will be asked the following questions to measure the similarity between haptic rendering and physical fabrics. }
		\label{question1}
        \scalebox{0.85}{
		\begin{tabular}{l| l }
			\hline
			\multirow{2}{*}{Q1} & \multirow{2}{*}{{\makecell[l]{Which of the showed three fabric pieces does  \\  the haptic rendering match with?}}} 
            \\&\multicolumn{1}{c}{}
            \\
            \hline
			\multirow{2}{*}{Q2} & \multirow{2}{*}{{\makecell[l]{Does the haptic rendering have the same\\slipperiness as the physical fabric?}}} 
            \\&\multicolumn{1}{c}{}
            \\
            \hline
		\multirow{2}{*}{Q3} & \multirow{2}{*}{{\makecell[l]{Does the haptic rendering have the same\\texture as the physical fabric?}}} 
            \\&\multicolumn{1}{c}{}
            \\
            \hline
		\multirow{2}{*}{Q4} & \multirow{2}{*}{{\makecell[l]{How much realism of haptic rendering do you feel\\compared with physical fabric?}}} 
            \\&\multicolumn{1}{c}{}
            \\
		\hline
		\end{tabular}
		}
\end{table}



\section{Experimental Results and Analysis}
\begin{figure}
	\centering
	\includegraphics[ width=0.75\columnwidth]{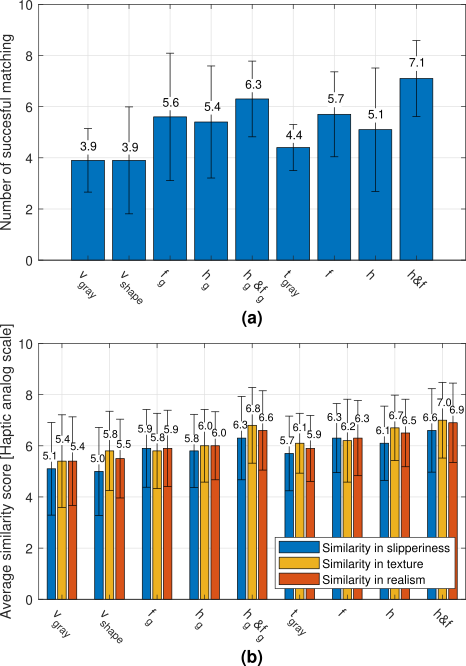}
	\caption{\textbf{Evaluation results from user study.} \textbf{(a)} average number of successful matching for each haptic rendering by different methods; \textbf{(b)} similarities of slipperiness, texture and realism by different methods. 
	}
	\label{fig:results}
	\vspace{-2.0em}
\end{figure}
The experimental results can be divided into three groups: (1) results of using the visual input for haptic rendering; (2) results of using the generated signals from visual images for haptic rendering; (3) results of using the collected signals from the tactile sensor for haptic rendering.
By comparing the results of these three groups, we would like to investigate the effects of various input modalities on haptic rendering as well as the effectiveness of our proposed methods.  


	

\begin{figure*}
	\centering
    \includegraphics[width=0.88\linewidth]{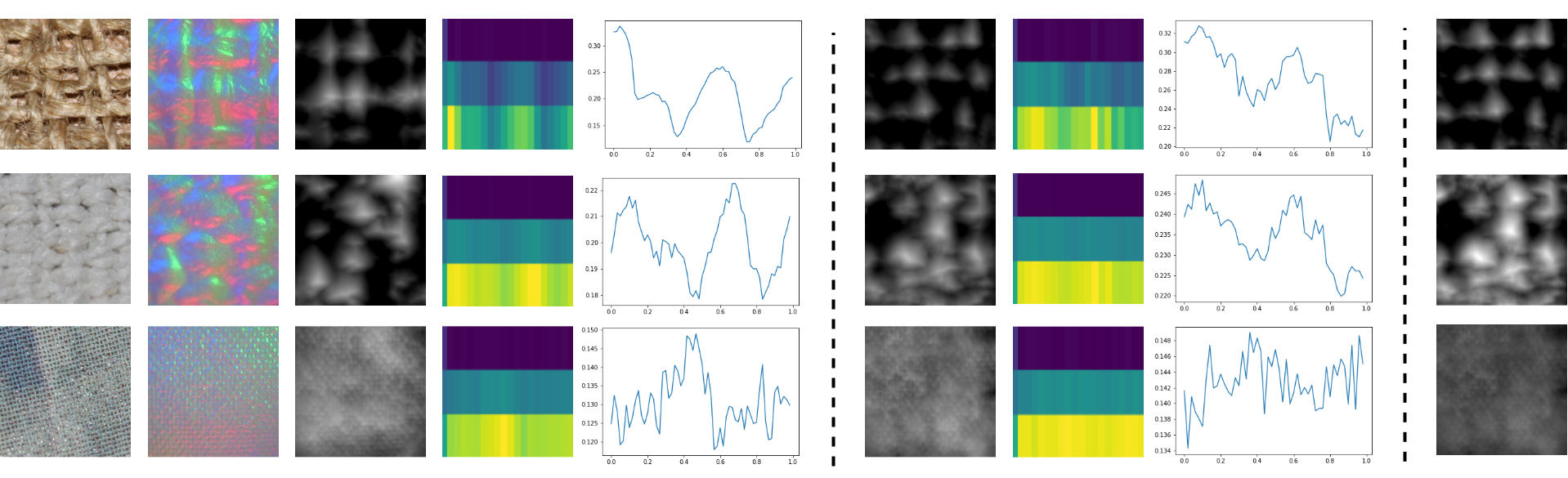}
	\caption{\textbf{Collected signals and generated signals from three example fabrics.} Left five columns: visual images; tactile images; height maps obtained from tactile images; spectrograms of friction coefficients;
	wave-format friction coefficients. Middle three columns: generated height maps; generated spectrograms; wave-format friction coefficients from generated spectrograms. Last column: generated friction images. }
	\label{fig:visualisation}
	\vspace{-1.8em}
\end{figure*}

\subsection{Do the vision-based methods work in haptic rendering?}

As can be seen in Fig.~\ref{fig:results}, each haptic rendering is only matched with the corresponding physical fabrics for 3.9 times on average, with $v_{grey}$ as input.
The similarity scores of slipperiness, texture, and overall realism are $5.1$, $5.4$, and $5.4$, respectively.
The use of $v_{shape}$, which extracts the height information from visual images, has a minor improvement in the similarities of texture and realism.
However, the overall performance is low, as the similarities of realism are around 5.5 out of 10 and the participants usually cannot match the rendering correctly in most cases. 
This could be due to that the tactile cues from visual images are limited and it shows that a simple mapping function for visual images cannot extract valid tactile information for haptic rendering.

\subsection{Do the height map and friction coefficients generated from vision improve the realism for haptic rendering?}
Compared with the methods using visual input, as can be seen in Fig.~\ref{fig:results}, the use of $f_{g}$ improves the haptic rendering significantly.
Each haptic rendering is matched correctly for 5.6 times on average, 1.7 times higher than the results of visual input.
Moreover, compared to the results using $v_{shape}$, the similarities of slipperiness and realism increase by 0.9 and 0.4 respectively. 
The reason could be that the generative model tries to reconstruct the friction information conditioned on visual images and friction coefficients during the training process, and the tactile cues can be preserved in the generated signals which leads to an improved result.

The generated height maps $h_{g}$, which contain the 2D texture geometry, improve the similarities in textures and realism by 0.2 and 0.1 respectively, compared to the results using $f_{g}$. 
In a further step, with $h_{g}$\&$f_{g}$ as input (Vis2Hap), the highest scores are achieved in all evaluation metrics among the methods using generated signals.
It means that the combination of $h_{g}$, which represents the degree of roughness, and $f_{g}$, which measures the slipperiness, is able to improve the realism of haptic rendering than using them separately.

\subsection{What is the difference between using collected signals and generated signals in haptic rendering?}

In the experiments that use collected signals in haptic rendering, it is observed that $t_{grey}$ results in the lowest performance as the mapping from colour tactile images to grey-scale images does not provide valid height or friction information for haptic rendering, as shown in Fig.~\ref{fig:results}. 
Results based on $f$, $h$ and $h$\&$f$ show a similar trend. However, they are higher than those based on generated signals.
Specifically, the use of $h$\&$f$ produces the best results among all experiments.
Each haptic rendering is matched correctly with the corresponding physical fabric for 7.1 times on average, and the similarity scores of slipperiness, texture, and realism are 6.6, 7.0, and 6.9 respectively.
It is worth noting that the results of our proposed Vis2Hap maintain at the same level compared to the results using $h$\&$f$.
Specifically, the average scores of similarities in slipperiness, textures, and realism are only 0.3, 0.2, and 0.3 less respectively, which demonstrates the effectiveness of our proposed method.

\subsection{Vision-based haptic rendering by cross-modal generation}
As shown in Fig.~\ref{fig:visualisation}, our generative model is able to generate realistic height maps and spectrograms from corresponding visual images. 
The generated results exhibit diversity and a high degree of similarity with the ground truth signals.
Moreover, it is observed that the intensity of the friction images changes compared to the height maps.
The changes in intensity are corresponding to the property of object's slipperiness to enhance the realism of haptic rendering.

For the quantitative evaluation of generated results, we use Mean Absolute Error (MAE) to measure the difference between generated friction coefficients and corresponding ground truth, and Structural Similarity Index Measure (SSIM)~\cite{wang2004image} to calculate the similarity between generated height maps and ground truth height maps.
The value of the MAE is 0.018, achieving the ratio of MAE and mean value of ground truth of 11.75\%, which means that generated friction coefficients and ground truth are similar.
The SSIM ranges from 0-1, where a higher value indicates a more identical result.
The SSIM between generated height maps and ground truth is 0.45.
Additionally, we perform a t-test to investigate whether generated friction coefficients and height maps have an identical average with the ground truth. Consequently, with a P-Value $<$ 0.05, our results are statistically significant, and the average of generated data and ground truth data are close. 
As a result, haptic rendering produced by our proposed method is comparable to approaches that use collected signals due to the high similarity between generated signals and ground truth.

\section{Conclusions}
In this paper, we propose a \textit{Vis2Hap} haptic rendering framework to generate the height maps and friction coefficients of the object's surface from vision, which are then applied for haptic rendering.
Based on the generated signals, two key characteristics of the object's surface, i.e., slipperiness and roughness, are combined to improve the realism of haptic rendering. 
Our Vis2Hap is capable of providing realistic haptic feedback of surface textures without requiring tactile data, reducing the workload of tactile data collection, and haptic rendering produced is comparable to approaches that use tactile signals. In the future, our Vis2Hap can be used to help human operators to assist robots to perceive objects and further facilitate robotic teleoperation. Our work also has a potential application in online shopping. For example, it is possible for people to feel the haptic properties of clothes without going to the shopping mall physically.






{\small
	\bibliographystyle{ieeetr}
	\bibliography{reference.bib}
}

\end{document}